%% file: main.tex
\documentclass[10pt,twocolumn,letterpaper]{article}

\usepackage{lipsum}
\usepackage{multirow}

\usepackage[pagenumbers]{cvpr} 

\input{preamble}

\definecolor{cvprblue}{rgb}{0.21,0.49,0.74}
\usepackage[pagebackref,breaklinks,colorlinks,citecolor=cvprblue]{hyperref}

\def\paperID{*****} 
\def\confName{3DV\xspace}
\def\confYear{2025\xspace}

\title{NeuroFly: A framework for whole-brain single neuron reconstruction}

\author{Rubin Zhao\\
{\tt\small rb.zhao@siat.ac.cn}
\and
Yang Liu\\
{\tt\small y.liu8@siat.ac.cn}
\and
Shiqi Zhang\\
{\tt\small sq.zhang1@siat.ac.cn}
\and
Zijian Yi\\
{\tt\small zj.yi1@siat.ac.cn}
\and
Yanyang Xiao\\
{\tt\small yy.xiao@siat.ac.cn}
\and
Fang Xu\\
{\tt\small fang.xu@siat.ac.cn}
\and
Yi Yang\\
{\tt\small y.yang@siat.ac.cn}
\and
Pengcheng Zhou$^{*}$\\
{\tt\small ­pc.zhou@siat.ac.cn}
}

\begin{document}
\maketitle
\input{sec/0_abstract}
\input{sec/1_intro}
\input{sec/2_related.tex}

\input{sec/3_pipeline.tex}

\input{sec/4_method.tex}

\input{sec/5_exp.tex}
\input{sec/6_con.tex}
{
    \small
    \bibliographystyle{ieeenat_fullname}
    \newpage
    \bibliography{main}
}

\end{document}

%% file: preamble.tex
\usepackage[dvipsnames]{xcolor}


%% file: sec/0_abstract.tex
\begin{abstract}
Neurons, with their elongated, tree-like dendritic and axonal structures, enable efficient signal integration and long-range communication across brain regions.
By reconstructing individual neurons' morphology, we can gain valuable insights into brain connectivity, revealing the structure basis of cognition, movement, and perception. 
Despite the accumulation of extensive 3D microscopic imaging data, progress has been considerably hindered by the absence of automated tools to streamline this process.
Here we introduce NeuroFly, a validated framework for large-scale automatic single neuron reconstruction. This framework breaks down the process into three distinct stages: segmentation, connection, and proofreading. In the segmentation stage, we perform automatic segmentation followed by skeletonization to generate over-segmented neuronal fragments without branches. During the connection stage, we use a 3D image-based path following approach to extend each fragment and connect it with other fragments of the same neuron. Finally, human annotators are required only to proofread the few unresolved positions. The first two stages of our process are clearly defined computer vision problems, and we have trained robust baseline models to solve them. We validated NeuroFly's efficiency using in-house datasets that include a variety of challenging scenarios, such as dense arborizations, weak axons, images with contamination.
We will release the datasets along with a suite of visualization and annotation tools for better reproducibility. Our goal is to foster collaboration among researchers to address the neuron reconstruction challenge, ultimately accelerating advancements in neuroscience research.
The dataset and code are available at \url{https://github.com/beanli161514/neurofly}.

\end{abstract}

%% file: sec/1_intro.tex
\section{Introduction}
\label{sec:intro}

\begin{figure}[t]
  \centering
  \includegraphics[width=\linewidth]{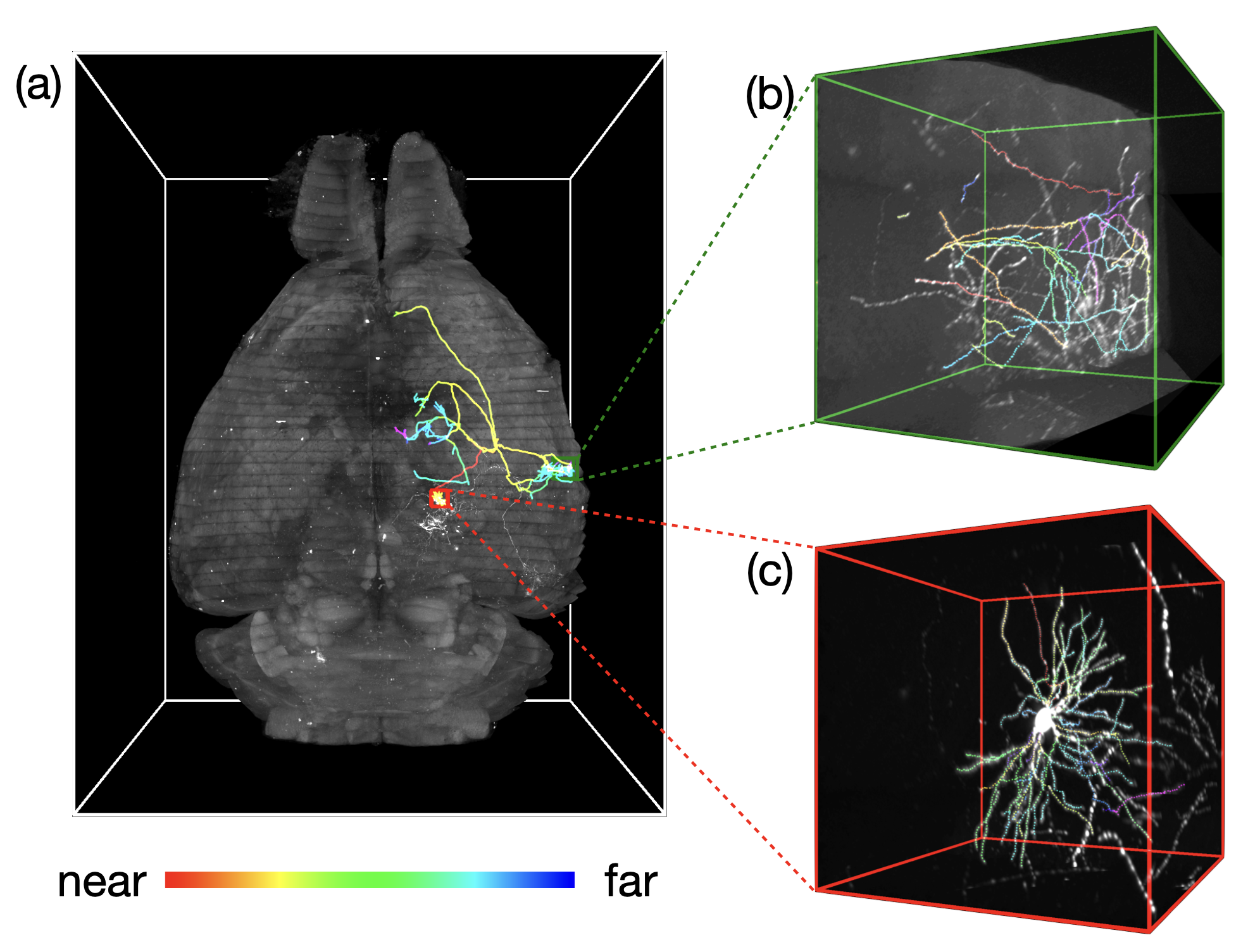}
  \caption{A mouse neuron reconstructed from fluorescence microscope images at a resolution of $1~\mu m/\text{voxel}$. (a) Image of a whole mouse brain and a reconstructed neuron. The color of the neuron segments indicates the structural distance, defined as the number of branch points encountered along the shortest path to the soma. The image size is $14000 \times 10000 \times 14000~\mu m^3$. (b) Arborization of the axon at the surface of the brain cortex, with an image size of $512 \times 512 \times 512~\mu m^3$. (c) Soma and dendrites of the neuron, with an image size of $384 \times 384 \times 384~\mu m^3$.}
  \label{fig1}
\end{figure}

Structural connectivity between neurons underpins communication across brain regions, forming the foundation for brain functions such as sentience and intelligence. Mesoscale imaging techniques \cite{xu2021highthroughput,zheng2013visualization,gong2013continuously} provide a powerful tool for studying the connections between different neuronal types across various brain regions \cite{zeng2018mesoscale}. Reconstructing neuron morphology from mesoscale brain images, particularly axonal projection, is crucial for building mesoscale connectome. Fig.\ref{fig1} (a) shows a mouse brain imaged at mesoscopic resolution. The neuron's axon begins at the soma as shown in Fig.\ref{fig1} (c), extends across thousands of voxels and spans to several brain regions. Manual annotation of such a mouse neuron without any automatic algorithm takes about 20 hours according to Gou \etal \cite{gou2024gapr}. Constructing a comprehensive mesoscale connectome for a species requires the reconstruction of tens of thousands of long-projection neurons, a task that is nearly impossible without advanced automated approaches.

Extracting neuron structures from the terabyte-scale images poses significant demands on the development of automated algorithms. The DIADEM challenge \cite[]{gillette2011diadem,brown2011diadem} was first launched in 2011 to benchmark and advance neuron reconstruction algorithms, later the BigNeuron project \cite[]{peng2015bigneuron} introduced standardized data protocols and evaluation methods to this field. As a result of these effort, many neuron reconstruction methods \cite[]{chen2021sphericalpatches,chen2022deeplearningbased,xiao2013app2,liu2018rivulet2,ming2013rapid} were published in the following years. However, recent benchmark \cite[]{Manubens2023BigNeuron} reveals that these neuron reconstruction algorithms struggle with complex scenarios not represented in existing datasets. It is obvious that a comprehensive, high-quality, and large-scale dataset is equally important as algorithms for advancing the construction of mesoscopic connectomes. Yet, annotating gold-standard datasets for developing and evaluating algorithms is as labor-intensive as labeling single neurons at the whole-brain scale. Therefore, a efficient data labeling approach is urgently needed.

In recent years, a variety of deep-learning based methods have been proposed for neuron reconstruction. Liu \etal \cite{liu2023tracing} trained a 3D segmentation network to improve the performance of traditional neuron reconstruction algorithms. Chen \etal \cite{chen2021sphericalpatches,chen2022deeplearningbased} combined spherical patches extration with a 2D CNN to predict the types and tracing directions of 3D critical points, significantly reducing computational cost. To avoid the high cost in obtaining manual annotations, Zhao \etal \cite{zhao2020neuronal} proposed a progressive learning scheme for neuron segmentation which learns from the coarse labels produced by traditional neuron tracing algorithms. However, whole-brain single neuron reconstruction was not solved end-to-end due to its complexity and low error-tolerance. To make deep-learng based methods more feasible in this field, a series of well-defined tasks with associated datasets is necessary.

The contributions of our paper are summarized as follows:

1. We introduce an efficient whole-brain neuron reconstruction framework, named NeuroFly, formulating whole-brain single neuron reconstruction as a streamlined process involving segmentation, connection, and proofreading. The segmentation and connection tasks are explicitly defined and supported by corresponding datasets.

2. We gather and label a variety of data of different species imaged using different techniques, constructing a diverse dataset covering a wide range of scenarios and varying scales up to whole mouse brain. The dataset is easily extendable benefiting from its succinct data protocol and the efficiency of NeuroFly framework.

3. We propose a 3D image-based path following method as part of the neuron reconstruction pipeline, addressing the gaps between neuron segments that were not correctly captured in the segmentation stage. This method serves as the baseline for the connection stage in NeuroFly framework.

%% file: sec/2_related.tex
\section{Related Work}
\label{sec:related}

\paragraph*{Curvilinear Structure Extraction.}
Curvilinear structures with intricate and elongated shapes are commonly found across various scales, including road networks, blood vessels, and nerve fibers. Various methods have been developed to extract these structures. Hand crafted feature extraction methods, such as Hessian-based filters \cite{steger1996extracting,botsaris1978curvilinear} detect curvilinear features by extracting regions with strong second-order derivatives. These techniques often struggle with challenges like contamination and scale variation. Deep-learning based methods \cite{cicek2016unet} performs well on such featrue extraction tasks but often require high-quality training data, which demands extensive human labor, especially for 3D data. Fortunately, Shit \etal \cite{shit2021cldice} proposed a differentiable skeletonization algorithm that enables end-to-end training of curvilinear structure segmentation models using only skeleton labels instead of mask labels. For neuron segmentation task, skeleton labels can be easily obtained by finding the brightest path \cite{jha2023brightest} between two manually labeled points using our annotation tool.

\paragraph*{Image-based Path Following.}
Image-based path following \cite{bojarski2016learning} is an essential technique in autonomous driving, which typically maps visual perception of current road environment to steering commands of the vehicle's control system. In some sense, curvilinear structures like nerve fibers can be analogous to tunnels in 3D space and the centerline serves as the target trajectory of a flying agent. This idea has been applied to electron microscopy (EM) image processing and has shown great performance in resolving split errors \cite{schmidt2024roboem}. In EM images, cell membrane of neurons can be clearly observed, so they are used as 3D lane markings and membrane-avoiding strategy was applied. Fluorescence images as showed in Fig.\ref{fig1} have lower resolution and larger scale comparing with EM images. The neurites appears as elongated lines with a width of about 3 voxels. So we train the agent to follow the neurite centerlines to achieve stable path following.

\begin{figure*}[t]
    \centering
    \includegraphics[width=\linewidth]{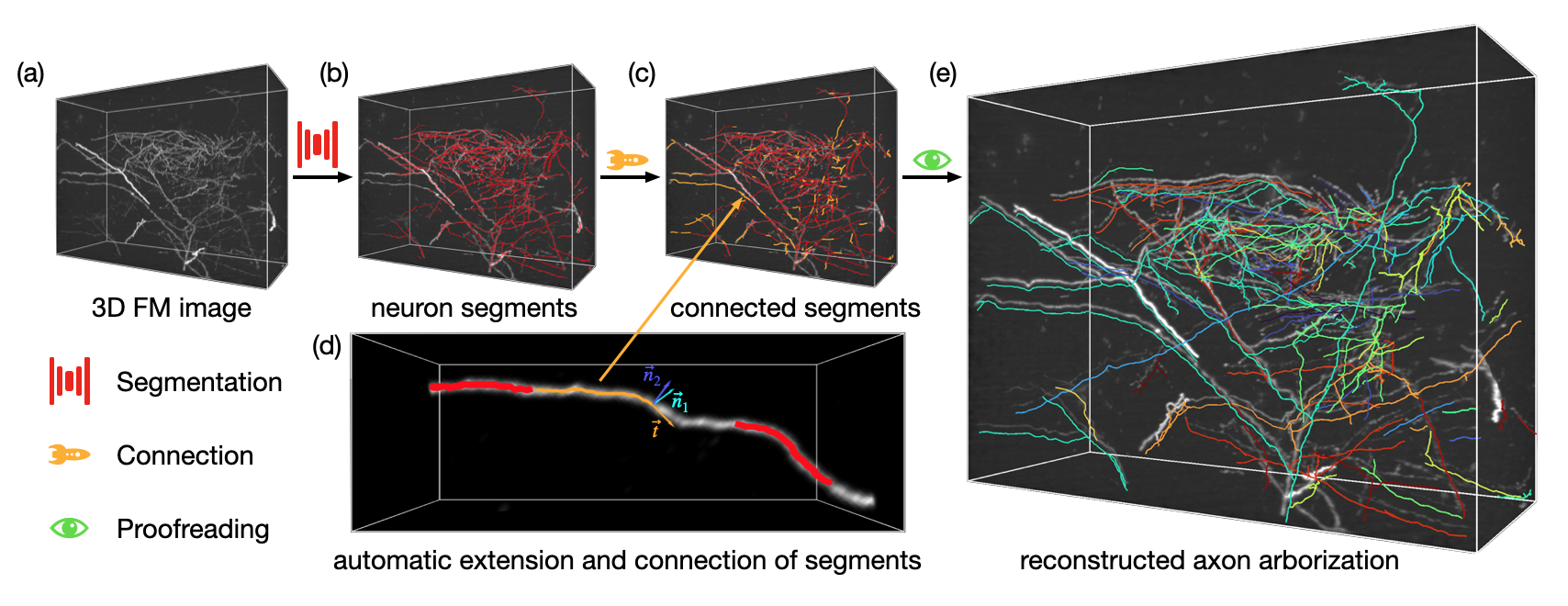}
    \caption{Illustration of NeuroFly's 3 stages. (a) Image of axon arborization used as example, with a size of $800 \times 600 \times 300 ~ \mu m^3$. (b) Result of the segmentation stage. Red lines represent neurites' centerlines extracted by the foreground segmentation followed by skeletonization. (c) Result of the connection stage. Orange lines represent the trajectories of agents starting off at the endpoints of each incomplete segments. (d) Kinetics and trajectory of the agent. $\vec{n}_1, \vec{n}_2, ~ \text{and} ~ \vec{t}~$ composes an rotation-minimizing frame, where $\vec{t}~$ denotes the tangent direction of the agent's movement. (e) Final reconstruction of the axon arborization. Neuron segments are shifted 10 voxels from the image and colored according to their connectivity for better visualization.}
    \label{fig:pipeline}
  \end{figure*}

\begin{table*}[t]
    \centering 
    \begin{tabular}{l c c c c c c c c c} 
    \toprule
    \multirow{3}{*}{Dataset} & \multicolumn{5}{c}{Contents} & \multicolumn{2}{c}{Label Types} & \multirow{2}{*}{Size}\\
    \cmidrule(l){2-6} \cmidrule(l){7-8} 
    & Dendrites & Axons & Arborizations & Indiv. Neurons & Contaminated & Dense & Sparse \\
    \cmidrule(l){2-9}
    Gold-166 \cite{peng2015bigneuron} & \checkmark & & & & & \checkmark & & MBs \\
    DIADEM \cite{gillette2011diadem} & \checkmark & \checkmark & \checkmark & & & \checkmark &  & GBs \\
    fMOST \cite{gong2013continuously} & \checkmark & \checkmark & \checkmark & \checkmark & \checkmark & & \checkmark & PBs \\
    WMBS \cite{chen2021sphericalpatches} & \checkmark & & & & & \checkmark & & MBs \\
    Ours & \checkmark & \checkmark & \checkmark & \checkmark & \checkmark & \checkmark & \checkmark & GBs\\
    \bottomrule
    \end{tabular}
    \caption{Overview of existing neuron reconstruction datasets.}
    \label{tab:datasets}
\end{table*}

\paragraph*{Semi-automatic Single Neuron Reconstruction.}
To get accurate morphology reconstruction of neurons, it is particularly important to ensure that all crucial locations like junction points and somas are connected correctly. A lot of software \cite[]{wang2022brainwide,longair2011simple,bria2016terafly,wang2019teravr,gou2024gapr} have been developed to increase the efficiency of human involvement in this task. 
Wang \etal \cite[]{wang2019teravr} applied virtual reality (VR) techniques to their annotation software, which greatly increases observability of dense scenarios. Gou \etal \cite[]{gou2024gapr} integrates automatic reconstruction and collaborative manual proofreading, allowing up to 100 annotators to collaborate on one dataset simultaneously. To make the reconstruction result more faithful, human annotators are required to proofread all the crucial locations. Besides, to label undetected neurites, annotators are required to click along them, which is extremely laborious. 



%% file: sec/3_pipeline.tex
\section{Task Formulation in NeuroFly Framework}
\label{sec:pipeline}

\begin{figure}[t]
  \centering
  \includegraphics[width=\linewidth]{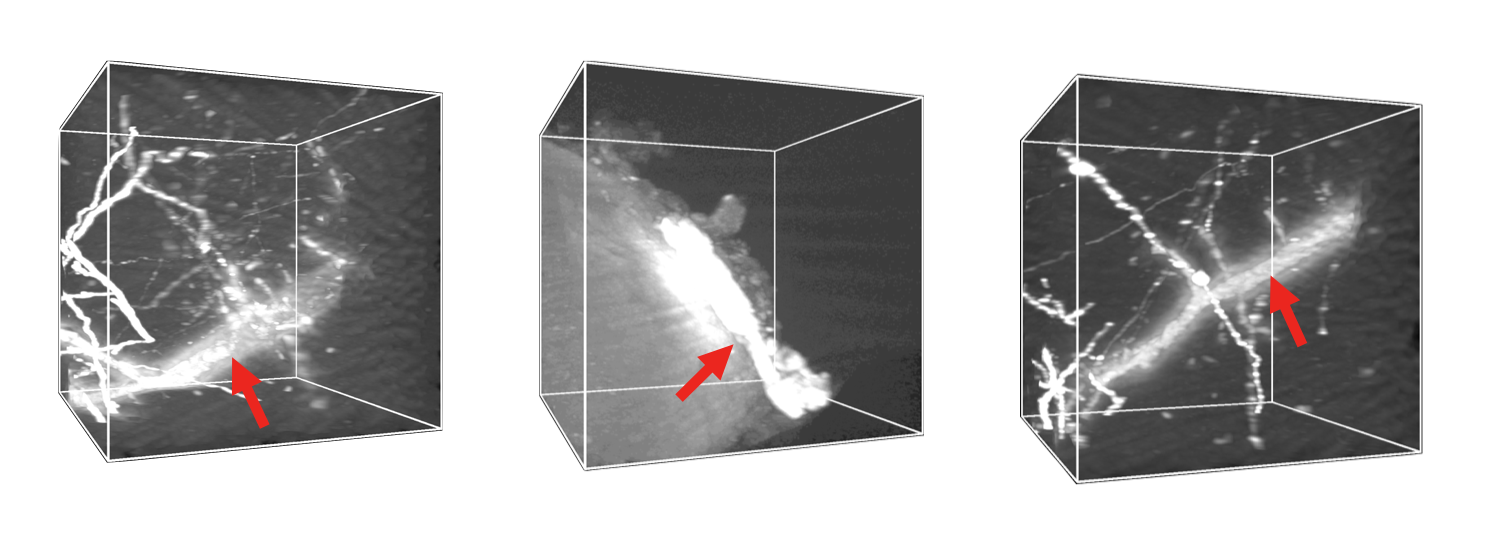}
  \caption{Examples of contaminated images, with contamination indicated by red arrows. The middle image shows a sheet-like matter adhering to the brain's surface that might be a piece of meninges. The other two images contain blood vessels.}
  \label{fig:nosiy}
\end{figure}

\begin{figure*}[t]
  \centering
  \includegraphics[width=\linewidth]{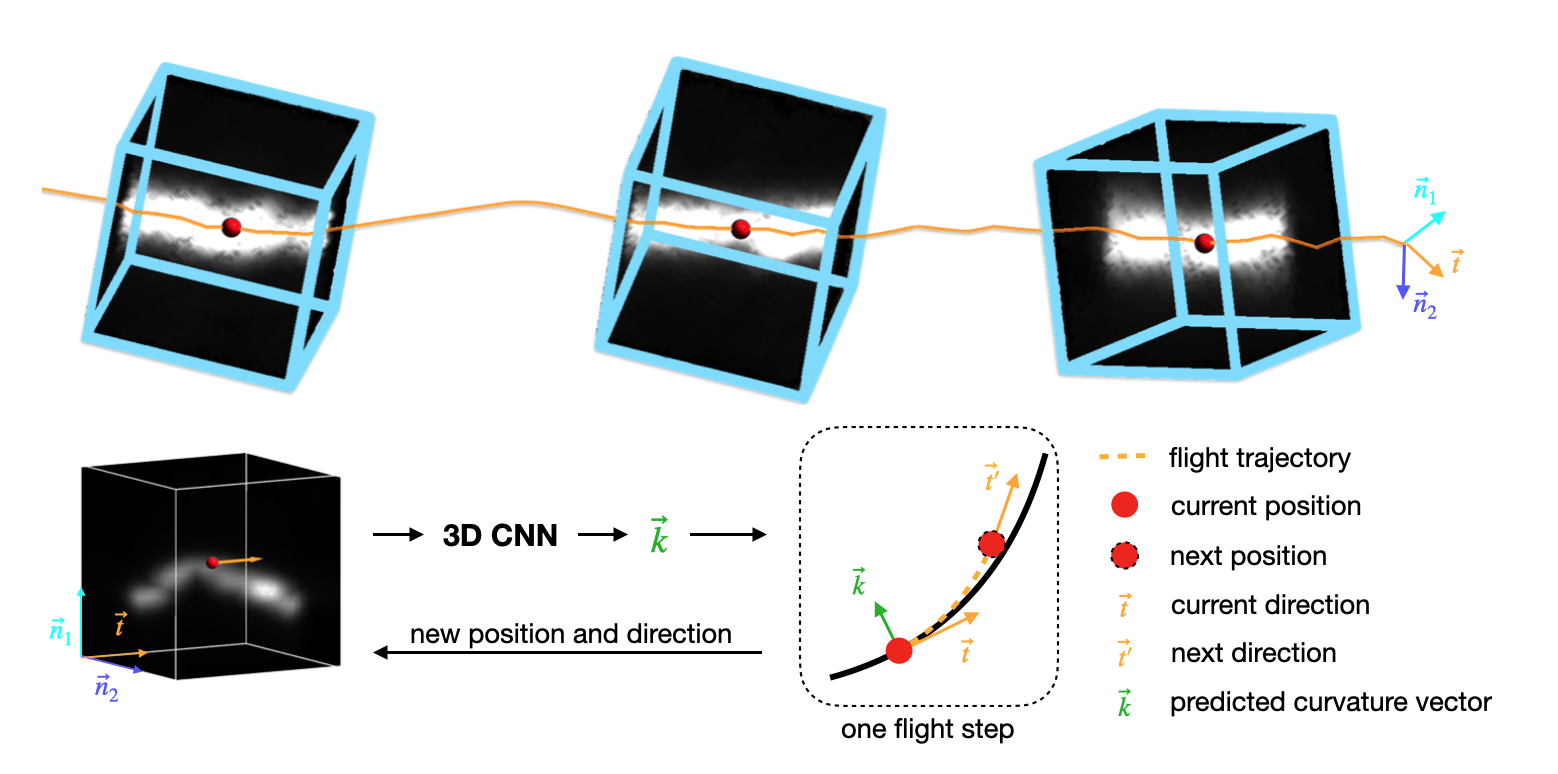}
  \caption{illustration of 3D image-based path following. The agent is denoted by red dot centering at its local image volume. Specifically, the directions of the x, y, and z axes of the local image align respectively with $\mathbf{n}_1$, $\mathbf{n}_2$, and $\mathbf{t}$.}
  \label{fig:driving illustration}
\end{figure*}

NeuroFly formulates neuron reconstruction task as a 3-stage streamlined workflow, composed of automatic segmentation, connection, and manual proofreading. Since deep-learning models are involved in the segmentation stage and connection stage, we carefully analyse the requirement of training data in various aspects to ensure the reproducibility of our method.

Here we demonstrate the process of whole brain neuron reconstruction from scratch using NeuroFly framework. 
Initially, when a new brain is imaged, it is recommended to label a tiny portion of it as training dataset for the segmentation model. Then with the trained model, image segmentation followed by skeletonization are perfromed on the entire brain image, that generates over-segmented neuronal fragments without branches as shown in Figure \ref{fig:pipeline} (b).
Then to extend and merge these incomplete segments, an agent start off one terminal of a segment and navigates along the neurite until it meets another segment or deviates from its path. This image-based path following is somewhat akin to autonomous driving in 3D space, as illustrated in Figure \ref{fig:pipeline} (d).
After these two stages, nearly all neurite centerlines are extracted and this result is sufficient for purposes like calculating neuron length, analysing neurite orientations \cite{li2023dlmbmap}, and training better segmentation models. However, to get precise morphology of neurons, manual proofreading of crucial locations is essential. A proofreading tool based on napari \cite{chiu2022napari} is also involved in NeuroFly framework, featuring succinct task design, intuitive interactivity and high-performance 3D rendering.


%% file: sec/4_method.tex
\section{Method}
\label{sec:method}

NeuroFly framework encompasses two major image analysis tasks. For the image segmentation task, we developed a novel data augmentation strategy that enables effective training with a very limited dataset. For the image-based path following task, we train a 3D CNN to predict the curvature vector of the parameterized neurite centerline, guiding the extension and connection of neuron segments.

\begin{figure}[t]
    \centering
    \includegraphics[width=\linewidth]{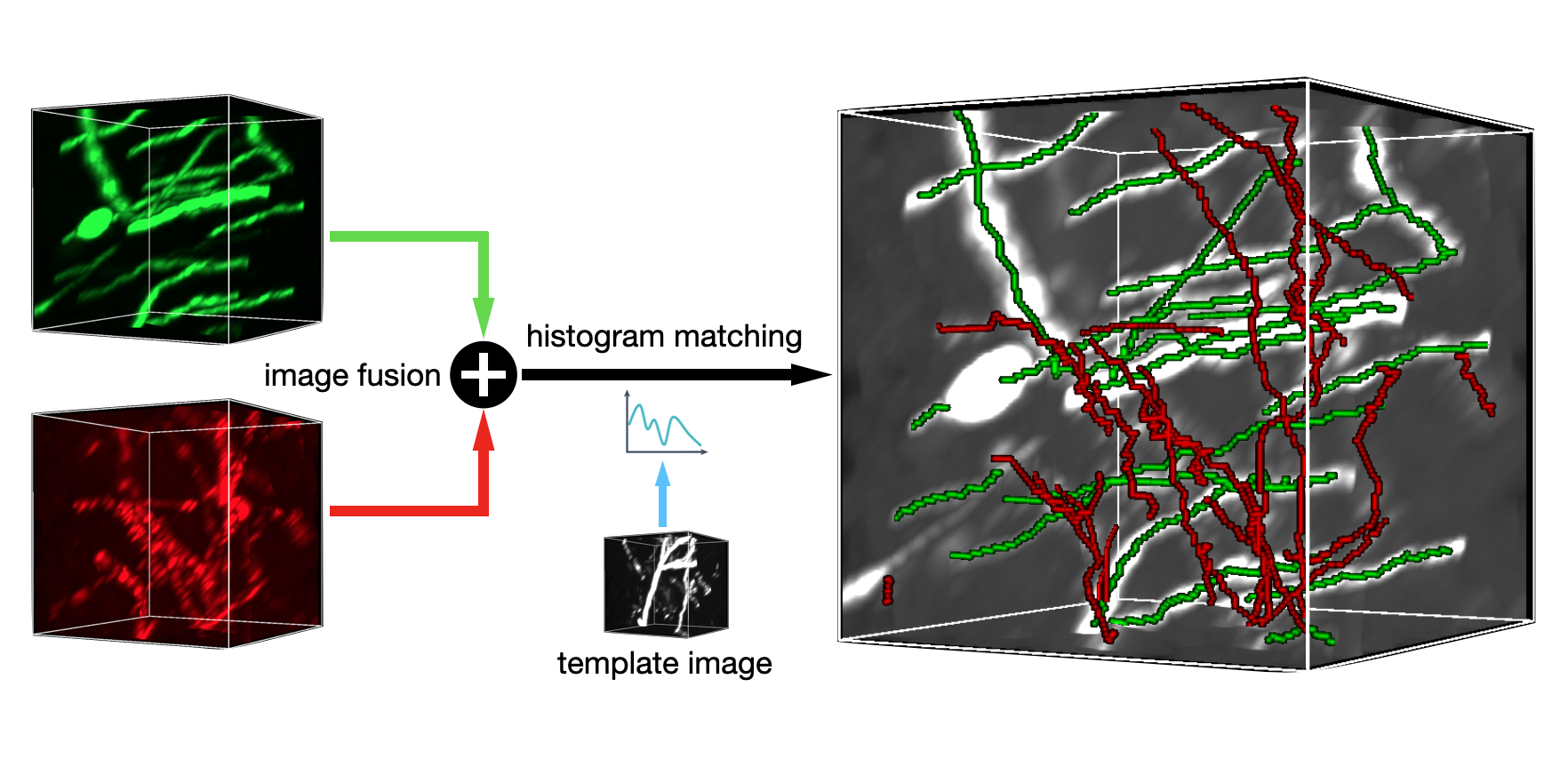}
    \caption{Illustration of data augmentation based on image fusion and histogram matching.}
    \label{fig:data augmentation}
\end{figure}

\subsection{Segmentation}

\paragraph*{Dataset Preparation} 
Many datasets have been released for a variety of purposes. Table. \ref{tab:datasets} compares our dataset with existing datasets based on the types of neuron structures they contain, the types of labels provided, and their sizes.
To train a model capable of segmenting all kinds of neuron structures in the inevitable presence of severe contamination, the training set must include densely labeled blocks that cover every scenarios in the entire brain image. Except common neuron structures illustrated in Figure \ref{fig1}, bright contamination including meninges and blood vessels as shown in Figure \ref{fig:nosiy} also exist ubiquitously across the brain. For each brain sample, we select 30 image blocks containing neuron structures, each with a size of $128^3$ voxels, along with some blocks containing only contamination. The skeletons of all neurites are labeled by finding the brightest path between two manually selected terminal points using our annotation tool. This data annotation process has been tested multiple times whenever we receive a new type of data, and through this, we have collected a large dataset comprising various types of data from different species, imaged using different techniques. 

\paragraph*{Data Augmentation}
As Figure \ref{fig:data augmentation} shows, the labeled samples are augmented using image fusion and histogram matching techniques to simulate images with varying luminance distributions and densely packed scenarios that are difficult to annotate. We prepare a training set comprised of 512 synthesized images and 512 contamination-only images gathered from 3 different brain samples.

\paragraph*{Model Configurations and Loss}
We use a vanilla 3D U-Net \cite{cicek2016unet} model with 3 layers consisting of 32, 64, and 128 $3 \times 3 \times 3$ kernels, respectively. Before processing the images with the model, min-max normalization is applied to map the 16-bit grayscale values to $[0,1]$ range. The model is trained using the clDice loss function \cite{shit2021cldice}, configured with 5 iterations of soft skeletonization.

\begin{figure*}[ht]
    \centering
    \includegraphics[width=\linewidth]{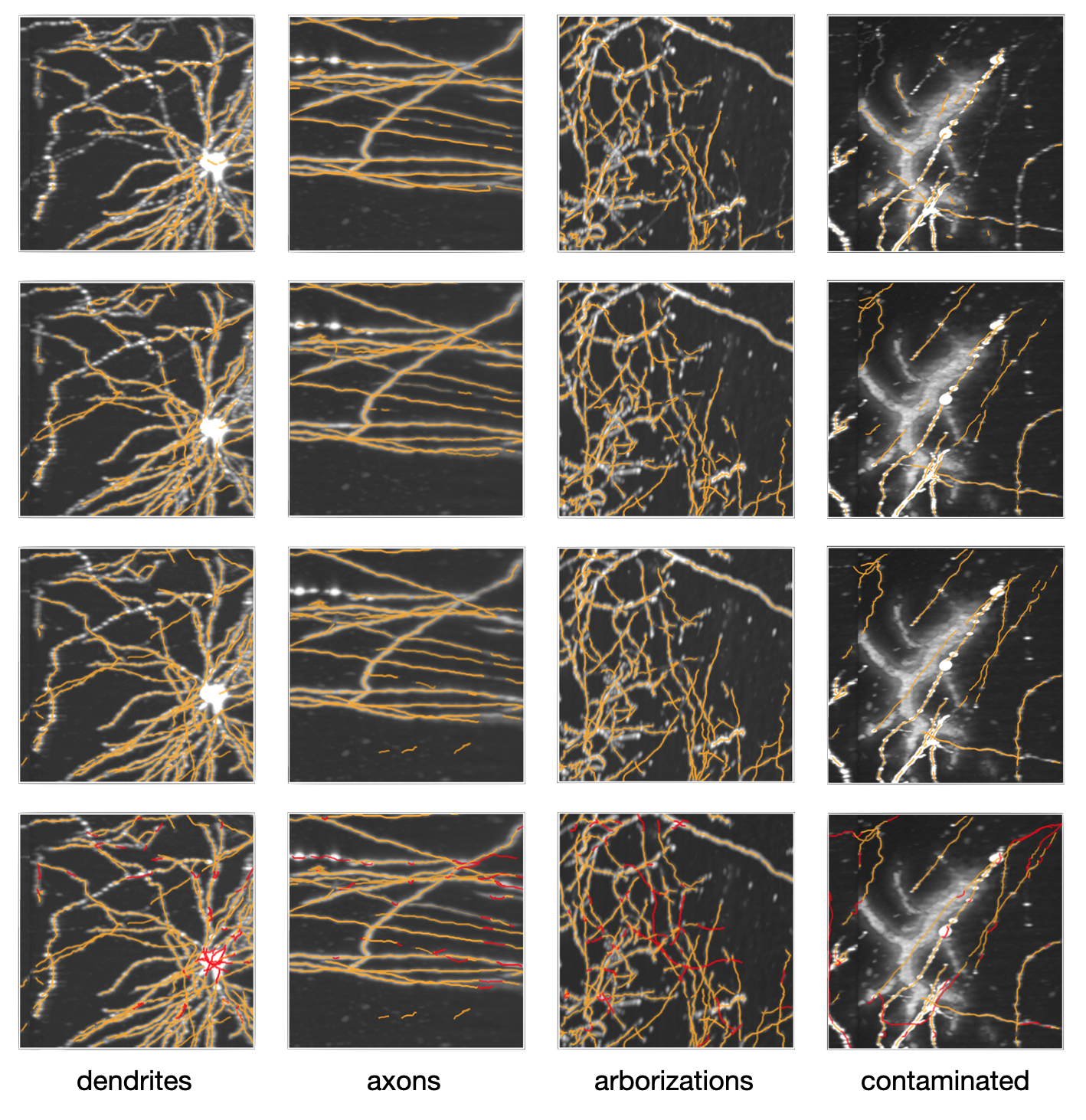}
    \caption{Comparison of methods on representative scenarios, including dendrites, axons, axon arborizations, and contaminated images. From top to bottom, each row displays the results of the following: MOST, segmentation model trained without data augmentation, segmentation stage of NeuroFly, segmentation stage followed by connection stage. Segmented skeletons are colored in orange, while trajectories generated during connection stage are colored in red.}
    \label{fig:result figure}
\end{figure*}

\begin{table*}[ht] 
    \centering 
    \begin{tabular}{l c c c c c c c c c c c c c c} 
    \toprule
    \multirow{3}{*}{Methods}
    & \multicolumn{3}{c}{Dendrites} & \multicolumn{3}{c}{Axons} & \multicolumn{3}{c}{Arborizations} & \multicolumn{3}{c}{Contaminated} & \multirow{2}{*}{Ave.} \\ 
    \cmidrule(l){2-4} \cmidrule(l){5-7} \cmidrule(l){8-10} \cmidrule(l){11-13}
     & rec. & prec. & F1 & rec. & prec. & F1  & rec. & prec. & F1 & rec. & prec. & F1 & F1\\
    \cmidrule(l){2-14}
    MOST & 0.505 & 0.969 & 0.664 & 0.819 & 0.926 & 0.869 & 0.787 & 0.884 & 0.833 & 0.571 & 0.361 & 0.429 & 0.763 \\
    M1 & 0.753 & 0.967 & 0.847 & 0.882 & 0.958 & 0.918 & 0.817 & 0.976 & \textbf{0.889} & 0.792 & 0.925 & 0.853 & 0.882 \\
    S1 & 0.849 & \textbf{0.981} & 0.910 & 0.858 & \textbf{0.987} & 0.918 & 0.715 & \textbf{0.987} & 0.829 & 0.904 & \textbf{0.982} & \textbf{0.941} & 0.898  \\
    S1+S2 & \textbf{0.882} & 0.974 & \textbf{0.926} & \textbf{0.961} & 0.931 & \textbf{0.946} & \textbf{0.950} & 0.810 & 0.874 & \textbf{0.938} & 0.609 & 0.738 & \textbf{0.913}  \\
    \bottomrule
    \end{tabular}
    \caption{Comparison of methods on representative scenarios. M1 refers to the segmentation model trained without data augmentation; S1 refers to the segmentation stage of NeuroFly; S2 represents the connection stage of NeuroFly. The weighted average F1 score is calculated by weighting the scores from different scenarios based on their length.}
    \label{tab:comparison}
\end{table*}

\paragraph*{Whole Brain Segmentation}
During inference, a threshold of 0.5 is applied to the network's output to generate a binary mask, which is then skeletonized to extract the neuron centerlines. To segment the whole brain image, we first chunk the terabyte-scale image into small blocks that fit into video memory. For this task, we use block sizes of $100 \times 100 \times 100$ voxels with 14-voxel borders on each side. After segmentation, the borders are removed from the output masks, and the small masks are concatenated in their original order. Next, skeletonization \cite{lee1994building} is applied to larger blocks of $300 \times 300 \times 300$ voxels. For brevity, we refer to segmentation and skeletonization simply as segmentation. The resulting skeletons are then sampled at specific intervals and saved as a graph. This entire process is highly parallelizable; with 8 RTX 4090 GPUs, segmentation of a whole mouse brain with a size of $14000 \times 10000 \times 14000$ voxels can be completed within 1 hour.

\subsection{3D Path Following}

Segmentation alone can not produce accurate neuron skeletons due to flaws introduced by the upstream processes, such as fluorescent tagging and imaging. For example, in Figure \ref{fig:pipeline} (d), the neurite appears much thicker than normal due to improper lighting and the convolution effect of point spread function (PSF) of the imaging system. As a result, a significant portion of this neurite is not correctly segmented during the segmentation stage. To address this problem, we introduce a connection stage, specifically designed to resolve false-negative results from the segmentation process. The connection process begins by deploying agents at the terminal points of each segment. The existing segments are splined to create a parameterized curve representation, which is then used to determine the initial position and orientation of the agents. With the initial condition set, the agent navigates along the neurite, guided by the steering signals predicted from the local image volume centering at and aligned with it. The agent follows the path until it merges with another segment or deviates to the background. This method ensures that most neurites not fully captured during the segmentation stage can be accurately connected, reducing the need for extensive manual annotation.

\paragraph*{Parameterization of Neuron Segments}
To describe the movement of the agent flying along the neurite centerlines, we parameterize the neurite skeletons and then construct a reference frame to describe the position and orientation of the agent. Here we apply degree 4 B-spline interpolation on the sparse skeleton points to get the parameterized curve $\mathbf{\mathbf{\gamma}}(t) \in \mathbf{C^3}$, with $t \in [0,1]$, where $\mathbf{\gamma}(0)$ and $\mathbf{\gamma}(1)$ represent the start and end point respectively. Unit tangent vector $\mathbf{T}$, unit normal vector $\mathbf{N}$, and curvature $\kappa$ can be calculated by:

\begin{equation}
\mathbf{T} = \frac{{\mathbf{\mathbf{\gamma}}}'}{\|{\mathbf{\mathbf{\gamma}}}'\|},
\end{equation}

\begin{equation}
    {\mathbf{N}} = \frac{{\mathbf{T}}'}{\|{\mathbf{T}}'\|},
\end{equation}

\begin{equation}
    {\mathbf{\kappa}} = \frac{{\mathbf{\mathbf{\gamma}}}'\times\mathbf{\mathbf{\gamma}}''}{\|{\mathbf{\mathbf{\gamma}}}'\|^3}.
\end{equation}

A segment may be straight in certain parts, resulting in zero curvatures. Since the Serret-Frenet frame requires non-zero curvature, we instead use the rotation minimizing frame (RMF) formed of unit tangent vector $\mathbf{T}$ and $\mathbf{N_1}$, $\mathbf{N_2}$:
\begin{equation}
    \mathbf{K}=\kappa\mathbf{N}=\kappa_1\mathbf{N_1}+\kappa_2\mathbf{N_2},
\end{equation}
where $\kappa_1$ and $\kappa_2$ are calculated using double reflection method \cite{wang2008computation}, and $\mathbf{K}$ is curvature vector.

\paragraph*{Curvature Vector Prediction}
Since the curvature vector determines the local continuation of a space curve, we predict it from the local image volume centered on and aligned with the neurite, as shown in Figure \ref{fig:driving illustration}. 
To achieve this, we employ a 3D convolutional neural network consisting of a series of residual blocks \cite{he2016deep}, followed by fully connected layers. The model transforms the local image volume into high-dimensional feature maps with Residual Blocks. These feature maps are then serialized into a 1D sequence through global average pooling and a flatten layer. The fully connected layers subsequently extract the desired curvature vector from this sequence. Notably, curvature vector lies within the plane spaned by $\mathbf{n}_1$ and $\mathbf{n}_2$, thus it has a constant value of zero in the dimension aligned with $\mathbf{t}$, so the model only predicts the 2 none-zero dimensions. While training, we calculate the mean squared error (MSE) between predicted vector and ground truth curvature vector defined as:
\begin{equation}
    \text{MSE} = \frac{1}{2} \sum_{i=1}^2(k_i-\hat{k}_i)^2 .
\end{equation}

\begin{figure}[t]
    \centering
    \includegraphics[width=\linewidth]{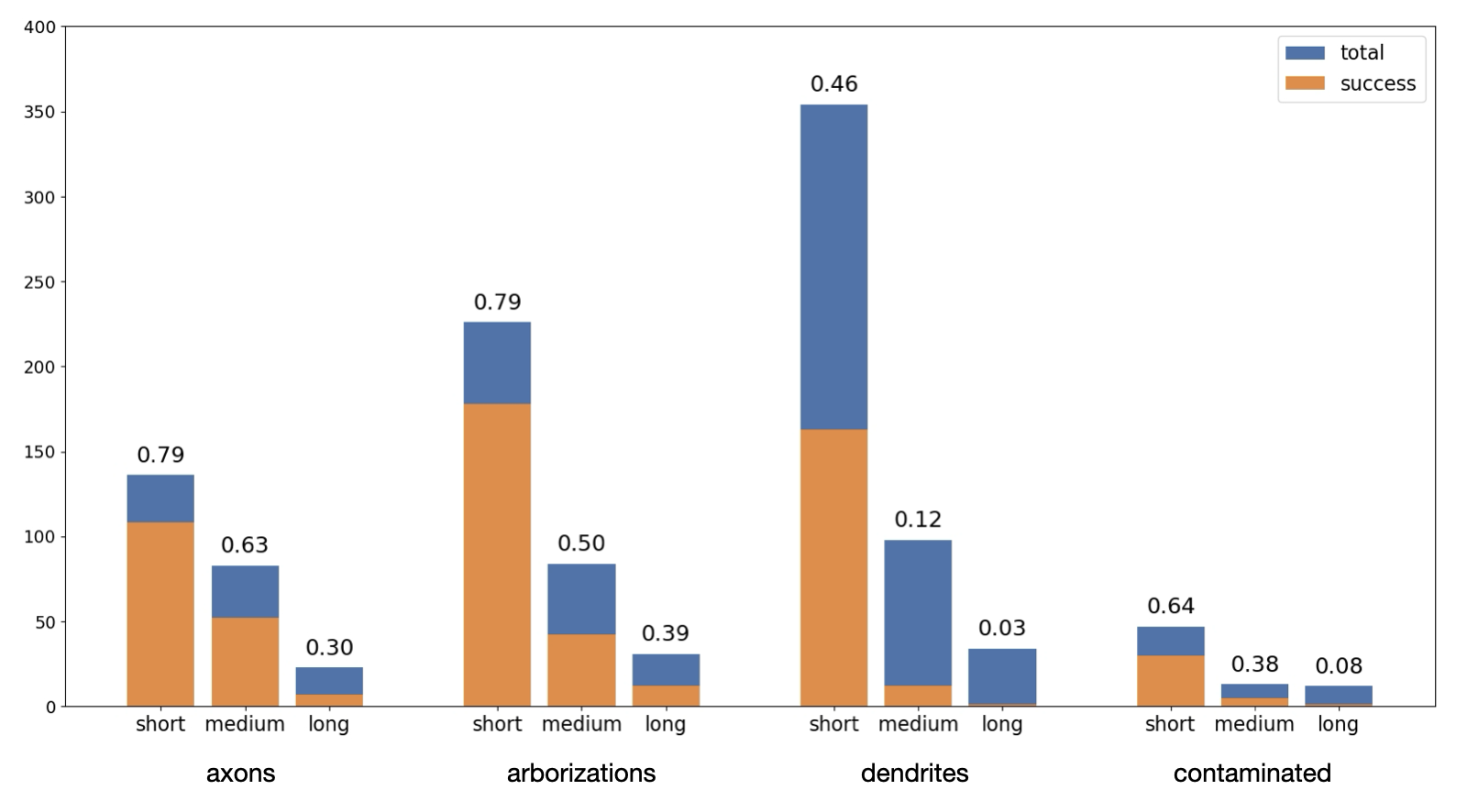}
    \caption{Stability of path following across different scenarios. The neurites are categorized into short, medium, and long based on their length distribution in each scenario. short: length $<~\mu$; medium: $\mu <= \text{length} < \mu + \sigma$; long: $\mu + \sigma <= \text{length}$. Here $\mu$ and $\sigma$ denote mean and standard deviation of the length distribution in each scenario.}
    \label{fig:connection results}
\end{figure}

\paragraph*{Dataset Creation} 
Training on samples extracted solely from neurite centerlines often results in poor performance during inference, as errors tend to accumulate with each step. To achieve stable path following, the training dataset must include off-centerline samples with corresponding curvatures that steer the agent back toward the centerline. To generate these samples, we shifted 20\% of the control points perpendicular to their tangent direction and then re-interpolated the curve to calculate the curvature required to pull them back. Additionally, 50\% of the images are fused with others to simulate more intricate scenes.

\paragraph*{Flying Strategy}
The degree 2 Taylor expansion of the arc-length parameterized trajectory curve $\tilde{\mathbf{\gamma}}$ 
\begin{equation}
    \tilde{\mathbf{\gamma}}(s)\approx \tilde{\mathbf{\gamma}}+s\tilde{\mathbf{\gamma}}' +\frac{s^2}2{\tilde{\mathbf{\gamma}}''}
    \label{eq:taylor}
\end{equation}
decomposes the motion of a point on the curve into the contributions of $\tilde{\mathbf{\gamma}}^{\prime}$ and $\tilde{\mathbf{\gamma}}^{\prime \prime}$, which are accordingly the unit tangent vector $\mathbf{T}$ and the curvature vector $\mathbf{K}$. The agent's future motion can be described as:

\begin{equation}
    \tilde{\mathbf{\gamma}}(s)\approx \tilde{\mathbf{\gamma}}+ s\mathbf{T}+\frac{s^2}2{\mathbf{K}},
    \label{eq:motion}
\end{equation}

\begin{equation}
    \mathbf{T}(s)\approx \mathbf{T}+s\mathbf{T'}=\mathbf{T}+s\mathbf{K}.
\end{equation}

Adaptive step size $\Delta s$ is used to control the agent's velocity, defined as:

\begin{equation}
    \Delta s=max(f\frac{d}{1+\frac{p}{2}\kappa},s_{\text{min}}) ,
\end{equation}

\noindent where $\kappa$ is curvature, $p$ is the physical size of the cropped image, $f$ is a step size factor, $d$ is an empirical value decided by the dataset, and $s_\text{min}$ is the minimal step size.


%% file: sec/5_exp.tex
\section{Experiments and Results}
\label{sec:exp}

\begin{figure*}[ht]
    \centering
    \includegraphics[width=1\linewidth]{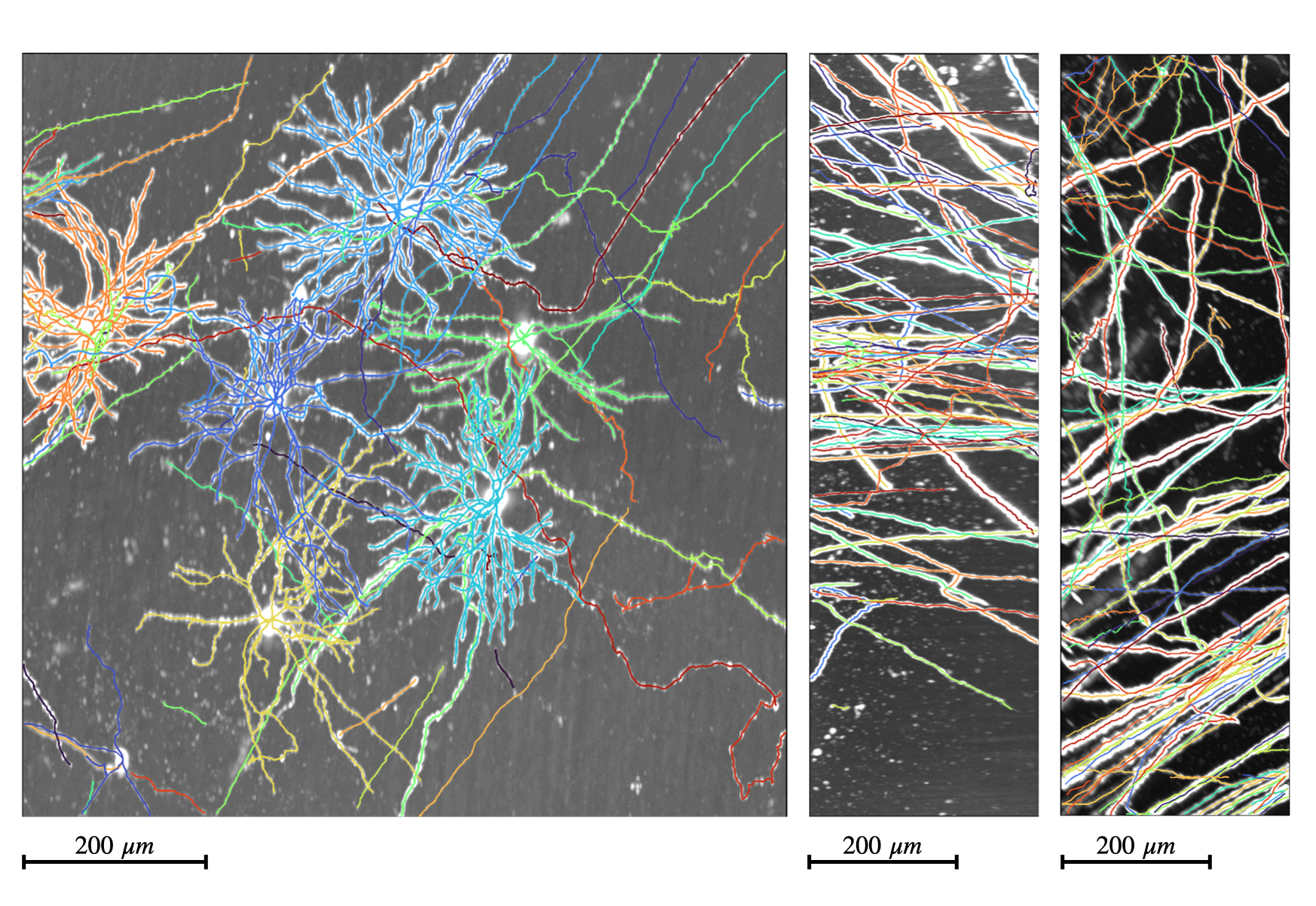}
    \caption{Per-instance labels produced by NeuroFly framework. The left figure shows the soma and dendrite regions of a mouse brain, while the middle and right figures display axons from a macaque brain.}
    \label{fig:final results}
\end{figure*}

\subsection{Implementation Details}
We train the curvature prediction model using the Adam optimizer for 1,000 epochs, with a 5-epoch warm-up. The initial learning rate is set to $10^{-4}$, and a step decay scheduler is applied. For the segmentation model, we use the Adam optimizer as well, training it for 200 epochs with a 10-epoch warm-up phase. The initial learning rate is set to $5 \times 10^{-4}$, and a cosine scheduler is used for decay. When predicting curvature, we crop a $32 \times 32 \times 32$ local image volume with a resolution of 1 $\mu m$/voxel as input. Therefore, $p$ is set to 16, and $f$, $d$, $s_{\text{min}}$ are set to 1, 2, 2, respectively.

\subsection{Comparison}
We tested our method with different configurations on four representative scenarios using data collected from multiple mouse brain and a macaque brain. For comparison, we also evaluation MOST \cite{ming2013rapid}, which is integrated in Vaa3D \cite{peng2014extensible}. As shown in Table \ref{tab:comparison}, the data augmentation method we propose enhances the segmentation model's performance in complex scenarios, such as dendrites and contaminated images. The connection stage we introduce significantly increases the recall rate across all scenarios with only a slight tradeoff in precision, resulting in an overall improvement of F1 score.

\subsection{Benchmark for Neurite Connection Task}
To quantify the performance of the image-based path following method used in the connection stage, we split neurons at junction points to create non-branching paths as the validation set. Two agents are deployed at both ends of each path and navigate toward the other end. The task is considered successful only if both agents successfully reach the opposite end. As shown in Figure \ref{fig:connection results}, our method works well only on simple tasks like axons and short paths. This benchmark provides standardized evaluation of neurite connection methods and can be also applied on any existing datasets.

\subsection{Results After Manual Proofreading}
Figure \ref{fig:final results} shows the per-instance labels produced by NeuroFly framework. The left figure contains six closely positioned neurons with intertwined dendrites. Accurate annotations of such intricate scenario provides valuable data sample for algorithm development and evaluation. With NeuroFly, these high-quality labeled datasets can be generated at scale, enabling the development of more advanced algorithms. As better algorithms emerge, they will facilitate the generation of even more refined data, creating a positive feedback loop that continuously enhances both data quality and algorithm performance.

%% file: sec/6_con.tex
\section{Conclusion}
\label{sec:con}

In this paper, we introduce NeuroFly, a framework for large-scale single neuron reconstruction. This framework formulates neuron reconstruction task as a 3-stage streamlined workflow, composed of automatic segmentation, connection, and manual proofreading. By standardizing each task, we establish a clear dataset protocol, enabling the continuous expansion of an extensible dataset. NeuroFly is designed to produce high-quality single neuron reconstructions, and it continuously improves as better algorithms are developed and larger datasets are incorporated. This makes NeuroFly a progressively evolving framework that enhances its capabilities over time. By proposing this framework, we aim to foster collaboration between the computer vision and neuroscience communities. 